\documentclass[final,12pt]{elsarticle}

\usepackage{graphicx,subcaption}
\usepackage{amssymb,amsmath}
\usepackage{cleveref}
\usepackage[utf8x]{inputenc}
\usepackage{todonotes}

\journal{Journal Name}

\begin{document}

\begin{frontmatter}


\title{An Average of the Human Ear Canal: Recovering Acoustical Properties via Shape Analysis}



\author[diku]{Sune Darkner}
\author[diku]{Stefan Sommer}
\author[BKSV]{Andreas Schuhmacher}
\author[BKSV]{Henrik Ingerslev}
\author[ZUH]{Anders O. Baandrup}
\author[ZUH,CUH]{Carsten Thomsen}
\author[BKSV]{Søren Jønsson}

\address[diku]{Department of Computer Science, University of Copenhagen}
\address[BKSV]{Br\"uel \& Kj\ae r Sound and Vibration}
\address[ZUH]{Zeeland University Hospital}
\address[CUH]{Copenhagen University Hospital}

\begin{abstract}
Humans are highly dependent on the ability to process audio in order to interact through conversation and navigate from sound. For this, the shape of the ear acts as a mechanical audio filter. The anatomy of the outer human ear canal to approximately 15-20 mm beyond the Tragus is well described because of its importance for customized hearing aid production. This is however not the case for the part of the ear canal that is embedded in the skull, until the typanic membrane. Due to the sensitivity of the outer ear, this part, referred to as the bony part, has only been described in a few population studies and only ex-vivo. We present a study of the entire ear canal including the bony part and the tympanic membrane. We form an average ear canal from a number of MRI scans using standard image registration methods. We show that the obtained representation is realistic in the sense that it has acoustical properties almost identical to a real ear.
\end{abstract}


\end{frontmatter}








\section{Introduction}
The anatomy of the outer human ear is well studied, in particular the Concha and the first part of the ear canal. However, due to an extremely sensitive surface inside the part of the ear canal which passes through the bone of the skull, the mastoid, traditional impression techniques can be very painful and the impression procedure itself may cause the subject to lose consciousness. This is due to the fact that the Auriculutemporal nerve of the parasympathetic nervous system passes close to the ear canal. Modern impression techniques have not developed beyond composite impressions used in connection with the fitting of hearing aids. This only enables us to record the geometry of the first part of the canal approximately 15-20 mm beyond the Tragus. The remaining part of the canal up untill the tympanic membrane, referred to as the bony part of the canal, has only been studied ex-vivo \cite{stinson1989specification}. Although attempts at scanning the outer ear have been made, most knowledge about the whole outer ear canal has been obtained ex-vivo \cite{stinson1989specification} and the shape analysis has only been performed manually. 

We set out to answer the following question: Is it possible to capture the geometry of the entire ear canal, and use shape analysis to generate a population average of the ear canal geometry that has acoustic properties comparable to real human ears? To answer this, we present a study and the underlying methodology for obtaining accurate measurements of the human ear using state-of-the-art scanning techniques: contrast enhanced magnetic resonance imaging (MRI). In contrast to previous in-vivo studies, this allows in-vivo capturing of the geometry of the outer ear canal from the Tragus and Anti-tragus to, and including, the typanic membrane. We use modern shape analysis techniques to create a population average which we validate by real physical impedance measurements, using specialized custom ear-moulds, on the same set of subjects as well as simulations. We show that the mean impedance is similar to the simulated impedance on the average geometry. This indicates that the average geometry has similar acoustical properties as that of the real human ears. We illustrate the anatomical variance of the average ear canal and show the average is a very good parametrisation of the population.

\section{Previous work}

The geometry of the human ear canal has received attention from the scientific community in the context of sound simulation and fitting of hearing and audio devices in the ear. One of the first in-depth investigations was conducted in \cite{stinson1989specification} where the shape of the human ear canal was investigated ex-vivo by making impressions on cadavers. The resulting impressions were then physically sliced in thin sections, and their shape described on a slice by slice basis. The extracted shapes were subsequently used for simulation in different scenarios \cite{stinson2005comparison,stinson2008sound}. An overview of acoustical models of the external/outer and middle ear can be found in \cite{rosowski1996models}. Sørensen et al. \cite{Sorensen2002a} made an image library of the human ear canal based on digital images of a cadaver, slice by slice in sub-millimeter resolution. One of the first in-vivo investigations of the shape of the ear-canal was published by Paulsen et al. \cite{paulsen2002building} who created the first artificial shape model of the human ear \cite{paulsen2003shape} based on the impressions of 29 subjects and expert annotations of the surfaces of the impressions. A more detailed model was presented by Darkner et al. \cite{darkner2007analysis} and extended to a model of the dynamic behavior of the ear canal \cite{darkner2008analysis}. Both papers used a fully automated alignment procedure \cite{darkner2008non,darkner06_reg}. 

 Further analysis and modeling of the shape of the ear canal has been carried out in \cite{unal2008customized,slabaugh20083,paulsen2004statistical} including attempts at building a model for use in automated production of hearing aids. Darkner et al. \cite{darkner2008analysis} used the shape of the ear and a penalized regression model to correlate shape changes due to ear motion with the perceived level of comfort by the subject wearing the hearing aid, however with inconclusive results.
 It has been been debated whether it is possible to locate the exact transition of the canal-bone junction based on ear impression alone. Here, Nielsen and Darkner \cite{nielsen2011cartilage} argued that the surface of the impression may reveal this junction by changes in the surface texture.
 
Shape models have been widely used in studies of biological shapes including brain morphology, exemplified by analysis of the human corpus callosum shape \cite{van2002active} and face morphology \cite{fathi_IMM2007-05250}. The Active Appearance Model (AAM) by Cootes et al. \cite{cootes_aam_bb36432} popularized linear models. In spite of the simplicity of linear models, they provide simple parametrisations of the shape space, and they have been found useful in many applications \cite{stegman_IMM2005-03355,larsen_darkner_IMM2007-04090,stegmann_thesisIMM2004-03126}. Linear models provide adequate representations of the nonlinear shape space when the variation in the shape data is limited. Because of the similarity of the shapes studied in this paper, we can therefore take advantage of the simple representation provided by linear shape models.

\section{Materials}
The data collected consists of acoustical measurements of the impedance of the ear canal together with the shape of the ear canal extracted from contrast enhanced MRI. The population consists of 44 adult Caucasian subjects, with normal hearing aged 24-55, with 12 female and 32 males subjects.
\subsection{Shape data}
The shapes were obtained from 44 contrast enhanced 3D T2-weighted MRI
axial scans of the human ear canal and surrounding tissue with a resolution
 of 0.22 x 0.22 mm and a slice thickness of 0.5 mm. The MRI scans (TR=1340
 ms, TE=260 ms) were acquired on a Siemens Avanto 1.5T scanner using a
surface coil. To enhance contrast, and subsequently the geometry of the ear
canal cavity, each subject was placed on the side and their ear canal was filled with rapeseed oil as a contrast agent.\cite{thomsen2017}. 
 
\begin{figure}[!h]
  \begin{center}
\includegraphics[width=0.92\linewidth]{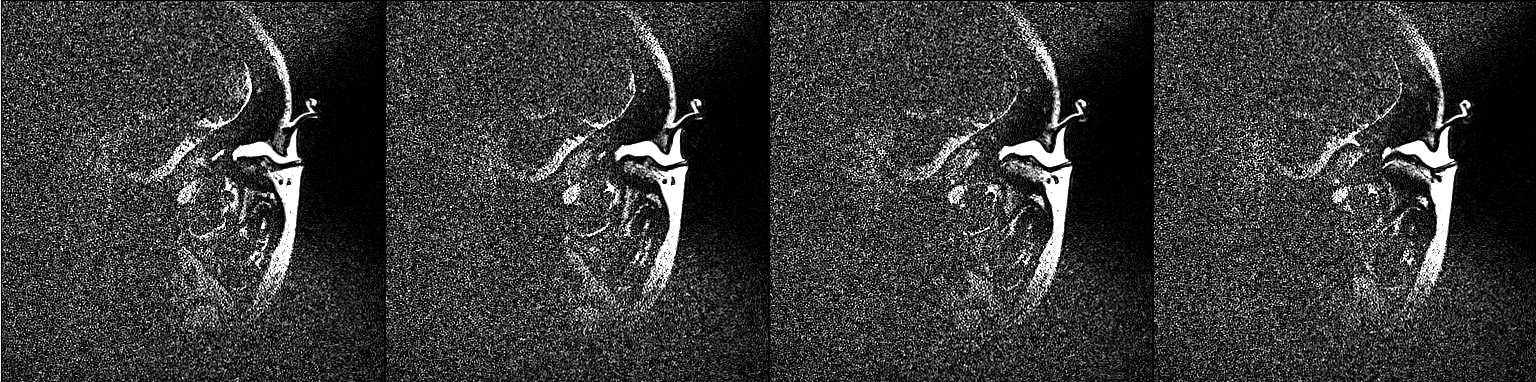}
\end{center}
\caption{\label{fig:t2} 4 consecutive slices in the transversal plane of the contrast enhanced T2-weighted MRI of the ear canal.}
\end{figure}

On a T2-weighted turbo spin echo sequence skin, muscle, bone, air and cartilage will appear dark and the only bright signal will come from fatty tissue such as subcutaneous fat and yellow bone marrow , see~\Cref{fig:t2}. Using oil as a contrast agent will prevent chemical shift artefacts and mis-registration. A T2-weighted pulse sequence was chosen because of the unpreceded high spatial resolution, and using oil as a contrast, gave a superior high signal to noise ratio. 
None of the volunteers felt any discomfort when their ear canals were filled with oil or after the scanning.
\Cref{fig:t2} shows 4 slices of the contrast enhanced ear canal.

\begin{figure}[!h]
  \begin{center}
\includegraphics[width=0.99\linewidth]{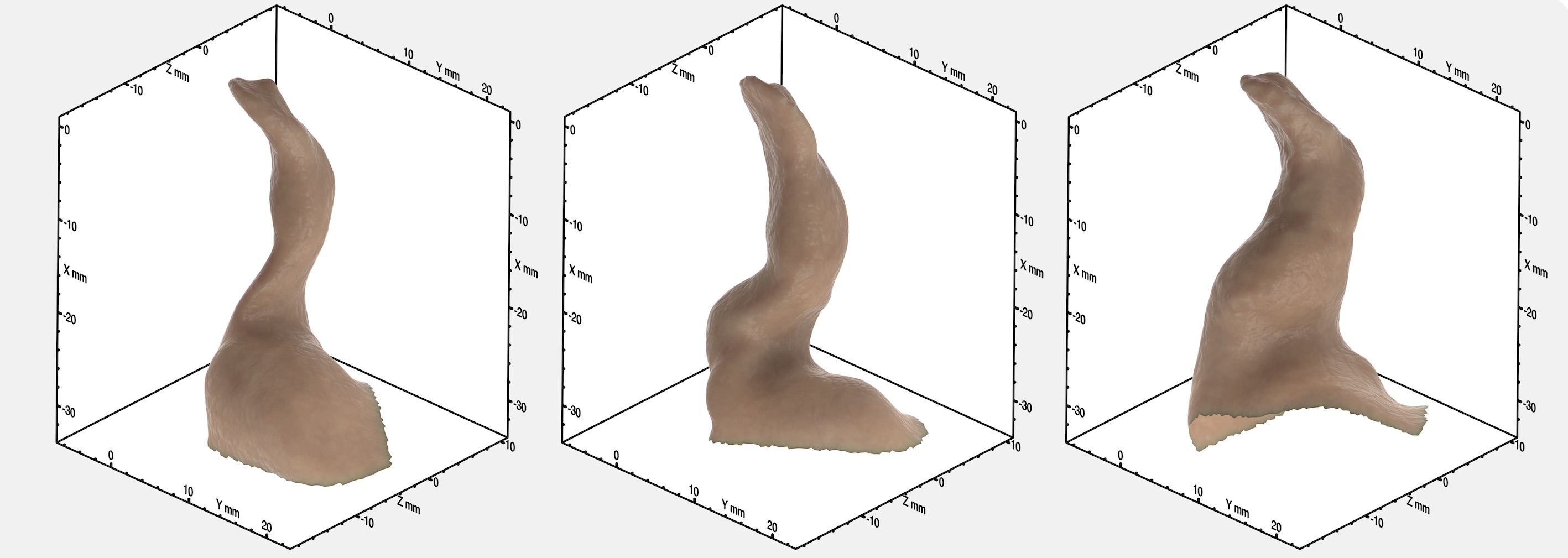}
\end{center}
\caption{\label{fig:surfs} 3 examples of the resulting segmented surfaces}
\end{figure}

\begin{figure}[!h]
  \begin{center}
\includegraphics[width=0.92\linewidth]{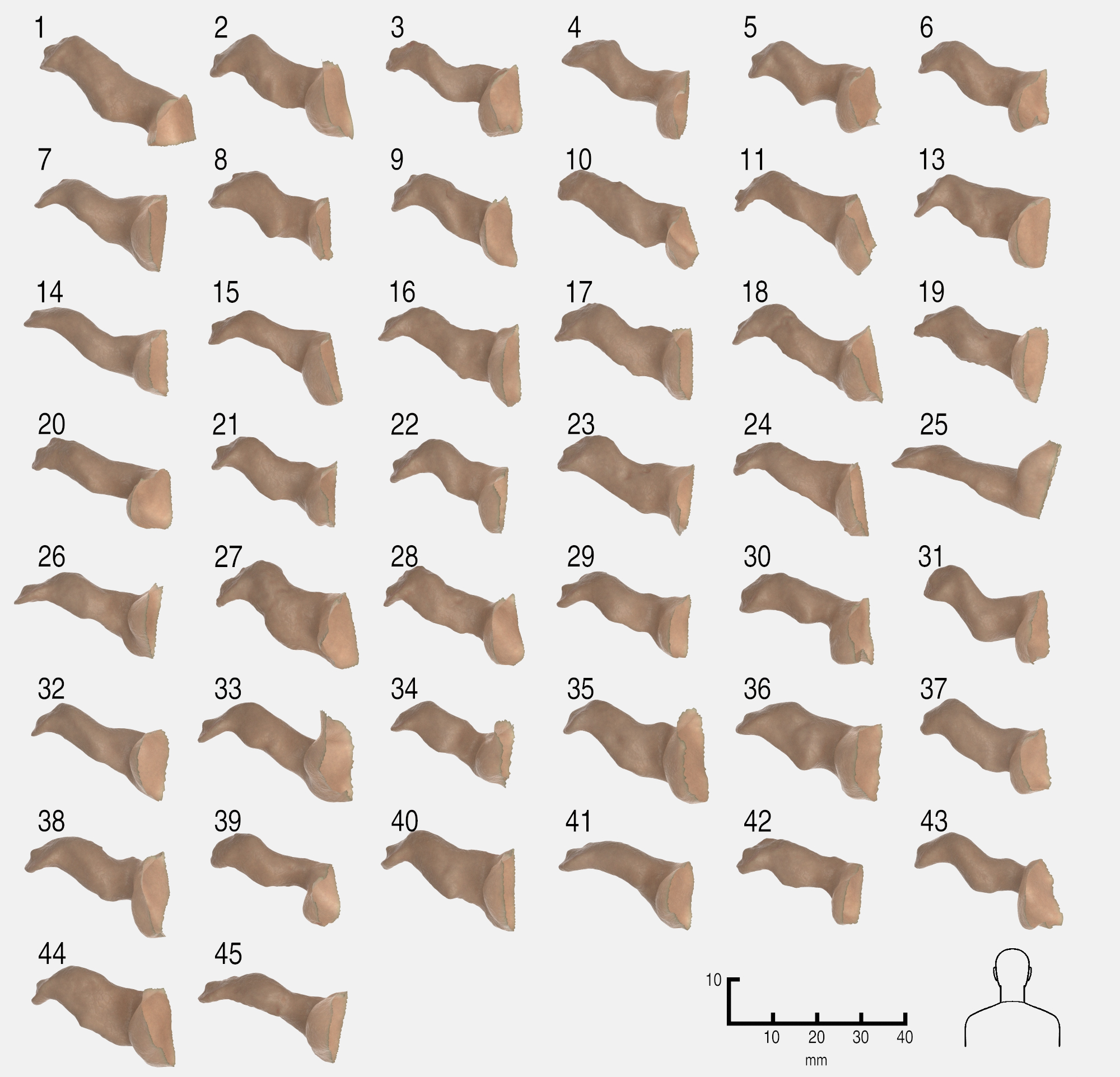}
\end{center}
\caption{\label{fig:surfs_all} The whole ear canal collection.}
\end{figure}

To recover the ear canal geometry from the image data, the MRIs were segmented manually by simple thresholding, and the surface was extracted from the binarized images using \cite{simpeware} to form a surface. The non-anatomical part of the surfaces were then removed manually such that the surfaces only represented the actual ear canal. The resulting surfaces are open 2D manifolds. 3 examples of the surfaces of the ear canals can be found in \Cref{fig:surfs} and an overview of all ear canal surfaces can be found in~\Cref{fig:surfs_all}.

\subsection{Acoustical Measurements}
Impedance measurements of the population were collected in the full audio band using a state of the art wide-band measurements technique described in \cite{jonnson2018jasa}.  
Using the impressions obtained from all the contrast enhanced MRI scans, a customized mould was produced for each individual subject, see~\Cref{fig:mold}. The mould was used to measure the acoustical impedance characteristics of the population as illustrated in \Cref{fig:mold}.  It may be difficult to obtain reliable acoustic measurements on human subjects even though careful preparations have been made. We discarded 12 measurements due to issues with repeatability and acoustic leakage (as explained below) and thus a total of 32 samples was included in the impedance calculations described below.

\subsubsection{Impedance measurements}
\begin{figure}[!h]
  \begin{center}
\includegraphics[width=0.99\linewidth]{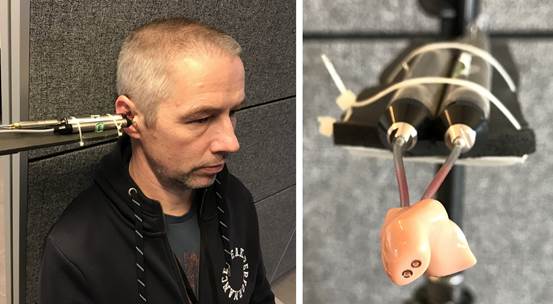}
\end{center}
\caption{\label{fig:mold} (a) A photo of the measurement set-up with the impedance probe assembly coupled to a real ear canal of one of the adult subjects. (b) A photo of the measurement set-up with one of the individual ear moulds made for all subjects connected to the impedance probe assembly through the two thin probe tubes.}
\end{figure}

A day or two prior to the measurements, the subjects had their ear canal carefully cleaned with an agent and lukewarm water to remove any unwanted residue such as wax. If a subject had a cold or flu  on the scheduled day of the measurement, the measurement was postponed. To obtain the measurement, the ear mould was inserted in the ear canal of the subject with the two probe tubes mounted, see \Cref{fig:mold}. The subject was then seated in a chair and the probe tubes were connected to the impedance probe assembly. A stepped sine sweep in 1/$24^{th}$-octave bands from 35 Hz to 25 kHz was then performed at a normal listening sound pressure level. All measurements were repeated until the same overall frequency response could be obtained ( three to four times). We ensured that no leakage was introduced between the ear mould and the ear canal, which usually appears as a roll-off at low frequencies.

\section{Methods}

To obtain the average ear canal geometry and to simulate the impedance for validation, the data was processed as described below.
\subsection{Surface registration and Average ear canal}

To form an average ear canal, we employed standard image registration techniques \cite{darkner06_reg,darkner2013locally,darkner2008non} to bring the surfaces of the ear canals into a common frame of reference. The problems of registrering two surfaces can be posed as follows: Given a surface $S$, the registration to a reference surface $R$ is estimated as:
\begin{align}
\mathrm{argmin}_\phi F(S,R,\phi)=M(S,R\circ \phi)+T(\phi)
\end{align}
where $\phi$ is the transformation that maps $R$ to $S$, $T(\phi)$ is a regularization of the produced transformation, and $M(S,R\circ \phi)$ is a similarity between the surface $S$ and the reference surface $R$ transformed by $\phi$.
All ears in the population are registered to the reference surface $R$, and $R$ is then iteratively updated to best represent the population. The surface to reference registration is performed first with an affine transformation \cite{darkner06_reg} and subsequently with non-rigid transformations. The updating of the reference surface follows \cite{darkner2007analysis}. The non-rigid transformations are represented using a B-spline FFD deformation model \cite{Rueckert1999a}, and the similarity between the transformed surfaces is measured using the locally orderless similarity measure \cite{darkner2013locally}. For this, the surfaces are represented as signed distance maps as described in \cite{darkner06_reg}, and the similarity measure is selected to be the $\ell_1$-norm between the signed distance maps. The registration processes follows the methodology presented in \cite{darkner2008non} where each individual shape is registered to the template, which is iteratively updated and repeated until convergence.

\begin{figure}[!h]
  \begin{center}
\includegraphics[trim={4cm 4.5cm 4cm 4cm},clip,width=0.82\linewidth]{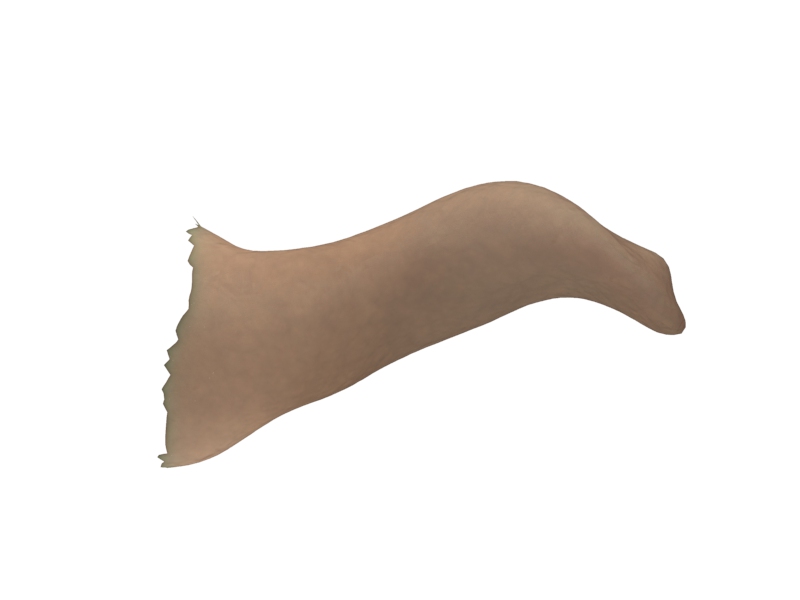}
\end{center}
\caption{\label{fig:mean} The average ear after registration}
\end{figure}

To construct the final parameterized average surface, we assume that the variation of the population is sufficiently small, such that a linear model of the shape variation is sufficient for representing the population and its variation. To this end, a mesh is selected and the vertices of the mesh are projected onto each of the individual surfaces in the template space. This will ensure that all points on the selected mesh (reference surface) have corresponding points on all other mesh surfaces in the population. From these correspondences, we form a point distribution model (PDM) with associated mesh and compute the mean shape~\Cref{fig:mean}.


\subsection{Average Impedance}
In order to obtain an average impedance of the population the measurements are propagated to a common reference plane in the ear canal using extended horn theory as described in~\cite{jonnson2018jasa2}. The  mean impedance is computed by correcting for the physical location of the measurement plane such that the acoustical distance to the tympanic membrane (the
half-wavelength resonance at 9.4 KHz) is identical across the population. At this plane the corrected responses are averaged to obtain the human average. For further details refer to~\cite{jonnson2018jasa2}.

\begin{figure}
\includegraphics[trim={2cm 0cm 2cm 0cm},clip,width=0.97\linewidth]{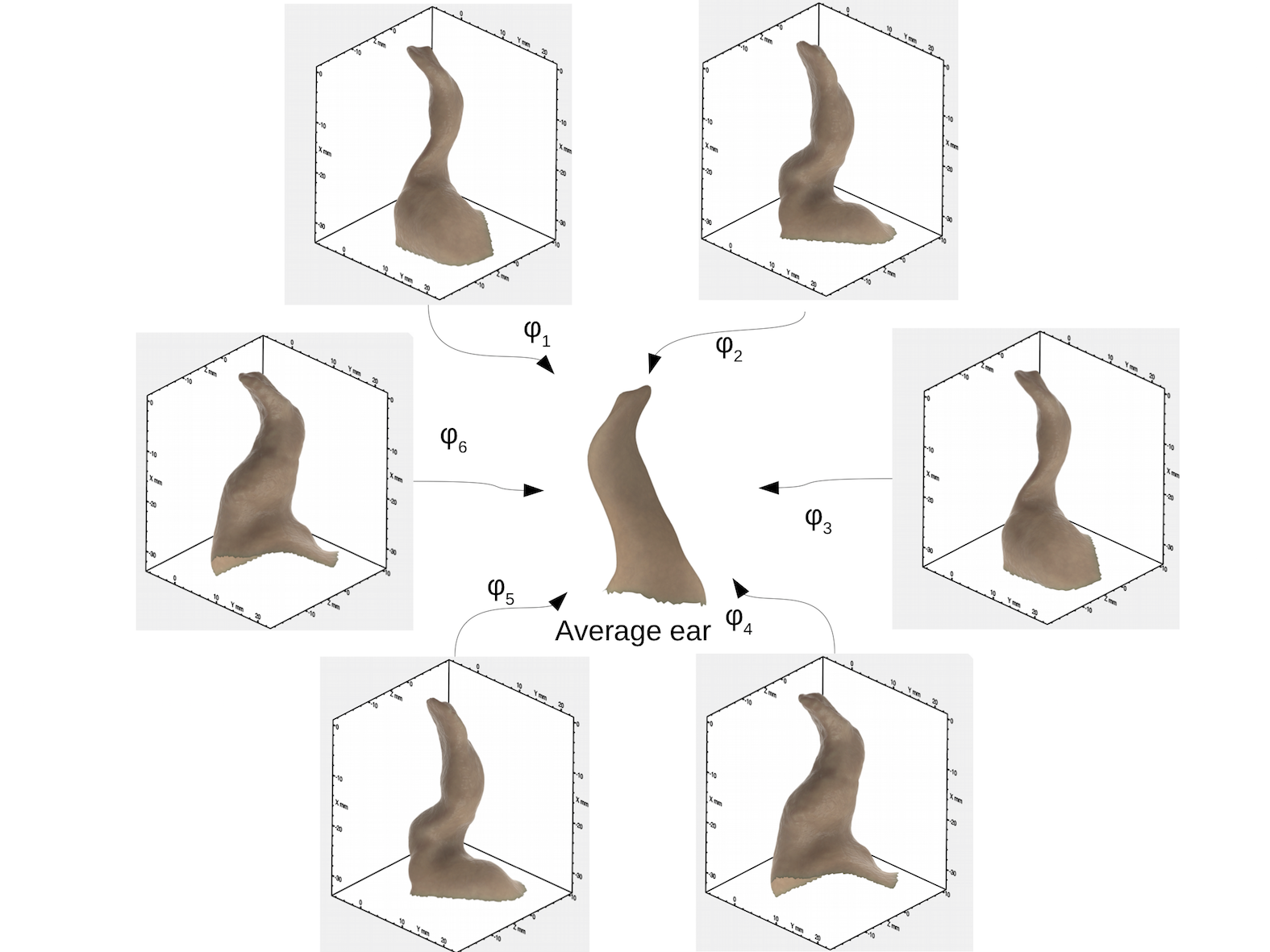}
\caption{\label{fig:registration}The figure illustrates a single step where the individual ears are registered to their common average, iteratively.  This is then updated which will form what is known as the Frechet mean in the space of deformations.}
\end{figure}

\subsection{Finite Element Modelling (FEM)}
\begin{figure}[!h]
\centering
\begin{subfigure}[t]{0.42\textwidth}
\centering
\includegraphics[width=1\textwidth]{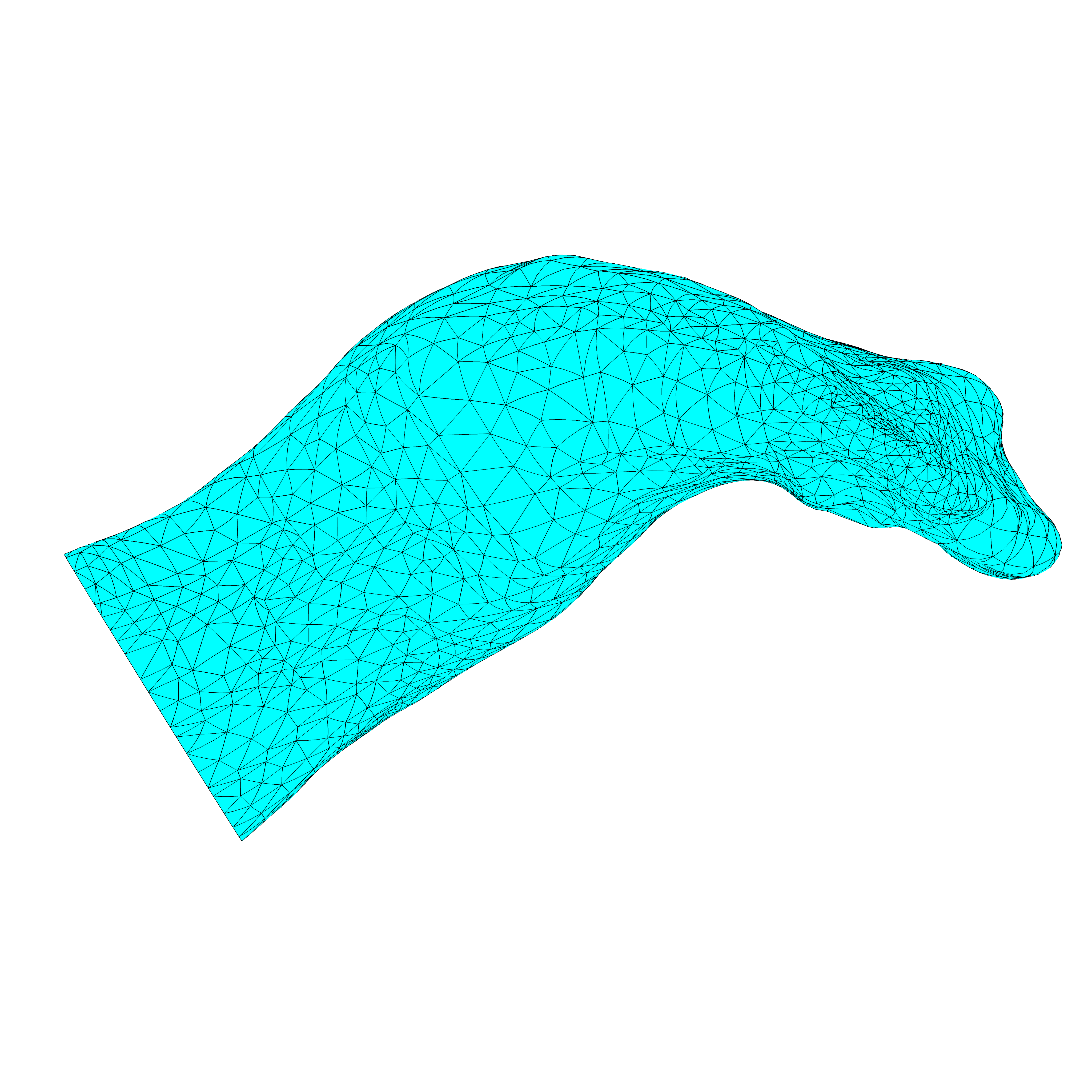}
\caption{\label{fig:mesh}Finite element mesh of real ear canal closest to average ear canal.}
\end{subfigure}
\hspace{1cm}
\begin{subfigure}[t]{0.42\textwidth}
\centering
\includegraphics[width=1\textwidth]{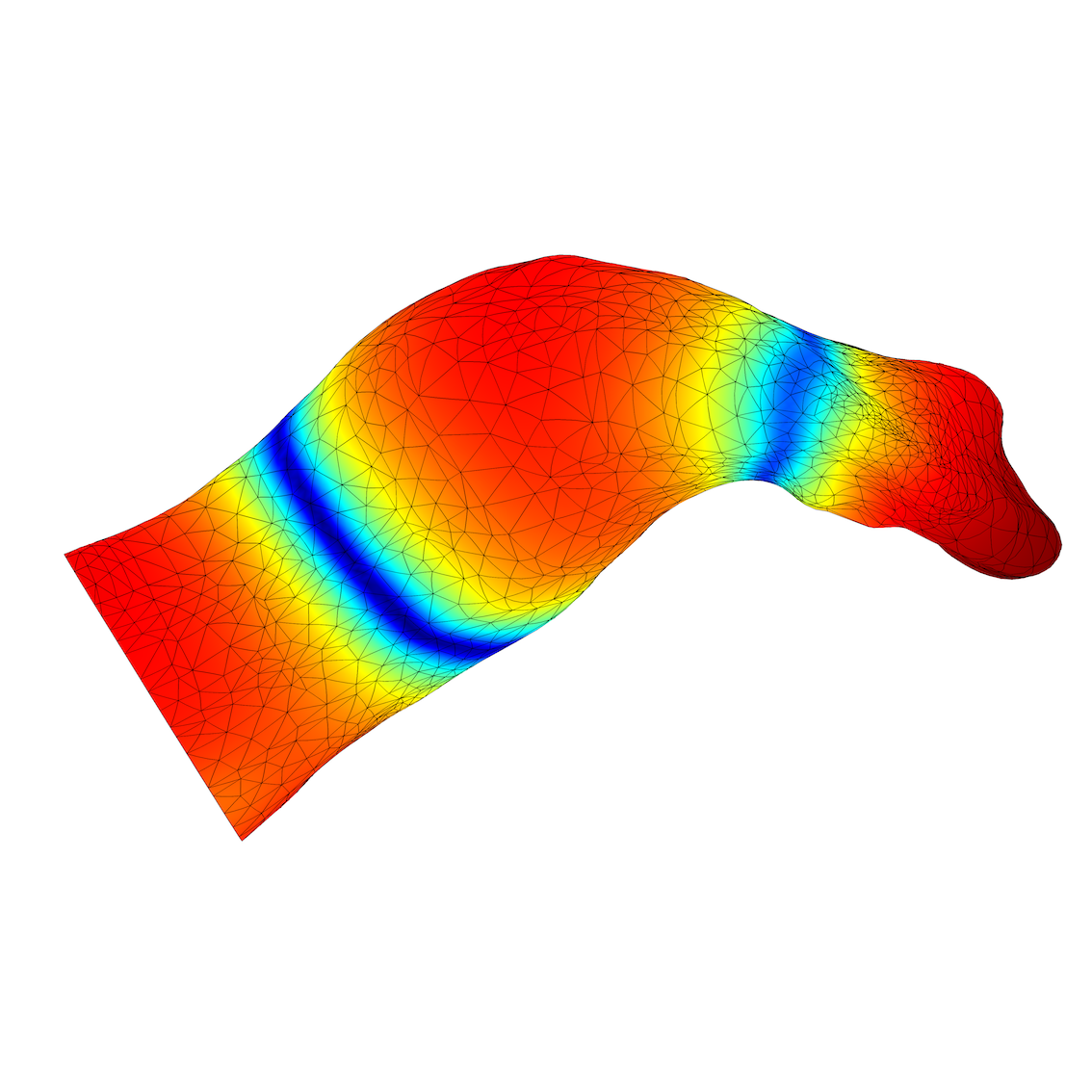}
\caption{\label{fig:spl}Simulated sound pressure distribution on ear canal boundary when exciting at the canal entrance.}
\end{subfigure}
\label{fig:sim}
\end{figure}

Simulations using the extracted ear canal geometries are performed using a FEM model. This is applicable because the problem is strictly to do with the interior of the ear canal, involving only the air domain inside each canal. 
The FEM was solved using COMSOL Multiphysics$^\text{\textregistered}$ Acoustics Module. To model the interior of the ear canal which consists of air, we use a volume mesh consisting of linear triangular elements. The effect of the eardrum impedance is imposed as a boundary condition at the end of the canal on a surface region representing the tympanic membrane and the remaining canal walls are considered as sound-hard. A small circular region at the canal entrance is acting as a piston exciting the air of the canal, similar to the physical ear canal measurements conducted on the individual subjects. 
We use a mesh density which at least resolves the acoustic wavelength with 6 elements, and thus the edge length of an element for a simulation up to 20 kHz should not exceed 3 mm, while matching the surface geometry of ear canal closely. The mesh in \Cref{fig:mesh} has a total of 15000 volume elements resulting in 23000 degrees of freedom to be solved for per specified frequency. After solving the problem with the given excitation acoustic quantities for example sound pressure distribution on the canal walls can plotted to understand the ear canal acoustics, see \Cref{fig:spl}. The ratio between the simulated acoustic pressure and excitation vibration at the canal entrance may be used to form the simulated acoustic input impedance. 

\section{Results}
The computed average ear canal, combined with our obtained measurements and the finite element simulation, allows us to verify the correctness of the average ear canal shape obtained from the ear geometry from the contrast enhanced MRI. 
\subsection{Shape}
\begin{figure}
  \begin{center}
  \centering
\begin{subfigure}[t]{0.32\textwidth}
\centering
\includegraphics[trim={4cm 4.5cm 4cm 5.2cm},clip,width=1\textwidth]{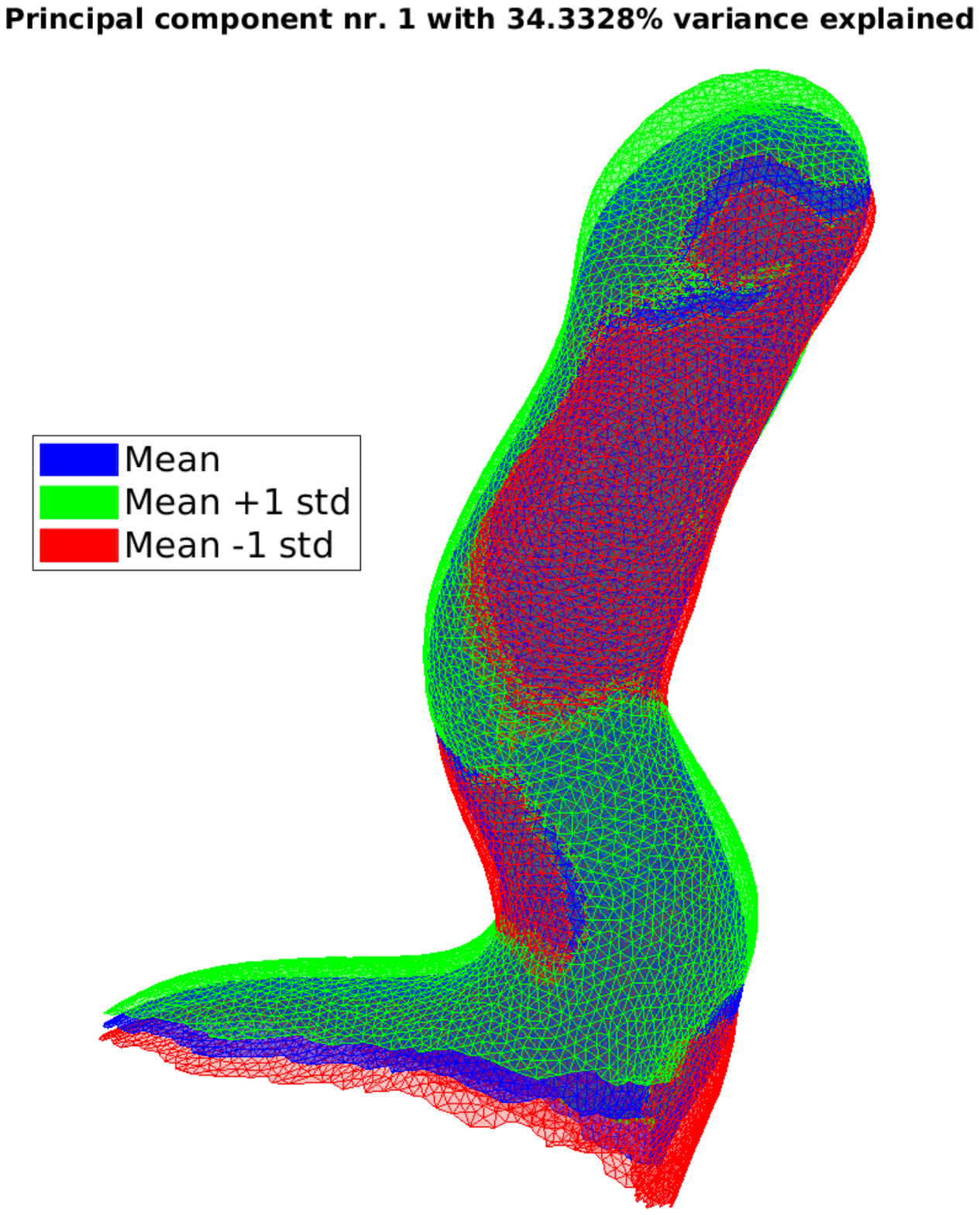}
\caption{PC1 +- 1 std.}
\end{subfigure}
\begin{subfigure}[t]{0.32\textwidth}
\centering
\includegraphics[trim={4cm 4.5cm 4cm 5.2cm},clip,width=1\textwidth]{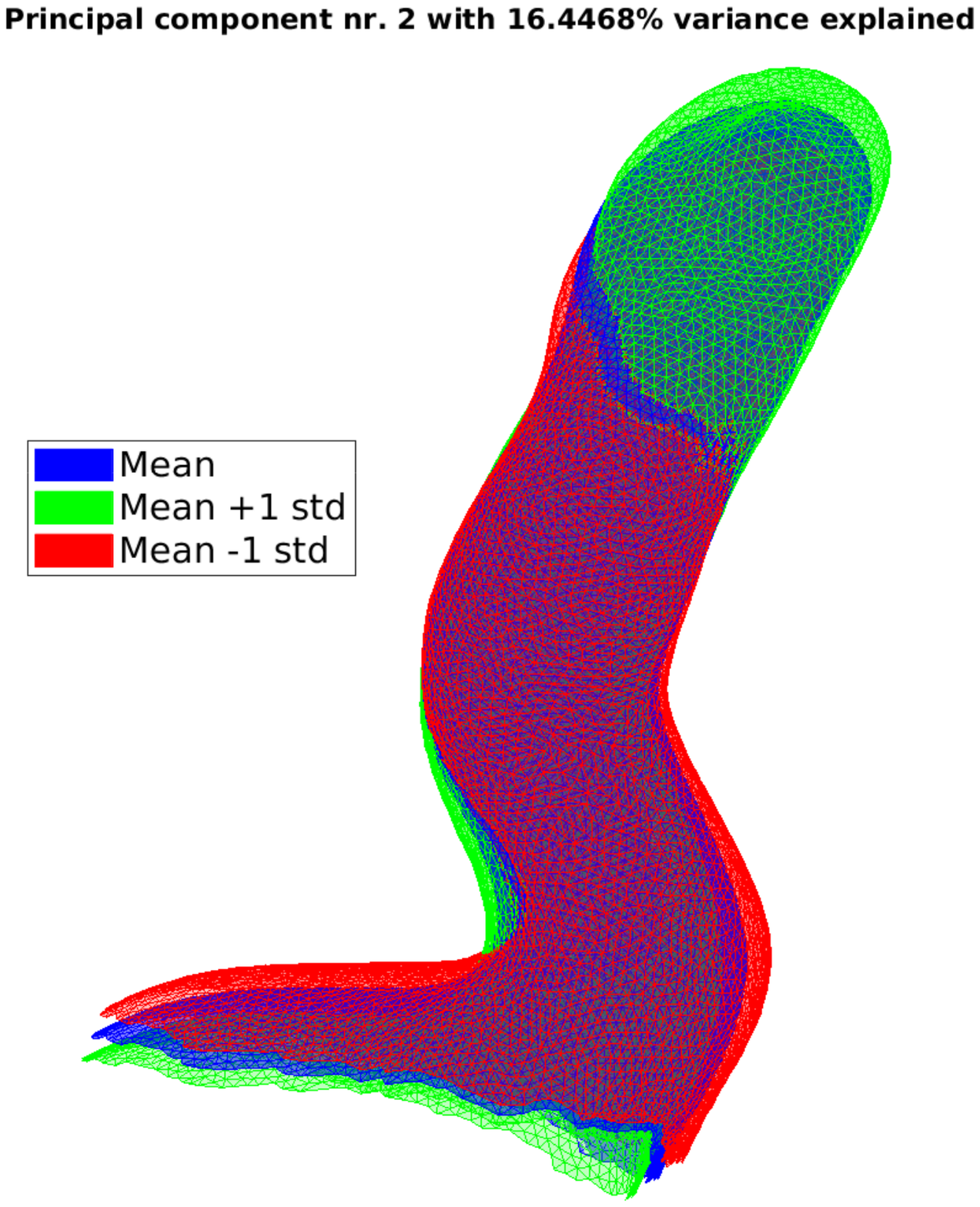}
\caption{PC2 +- 1 std.}
\end{subfigure}
\begin{subfigure}[t]{0.32\textwidth}
\centering
\includegraphics[trim={4cm 4.5cm 4cm 5.2cm},clip,width=1\textwidth]{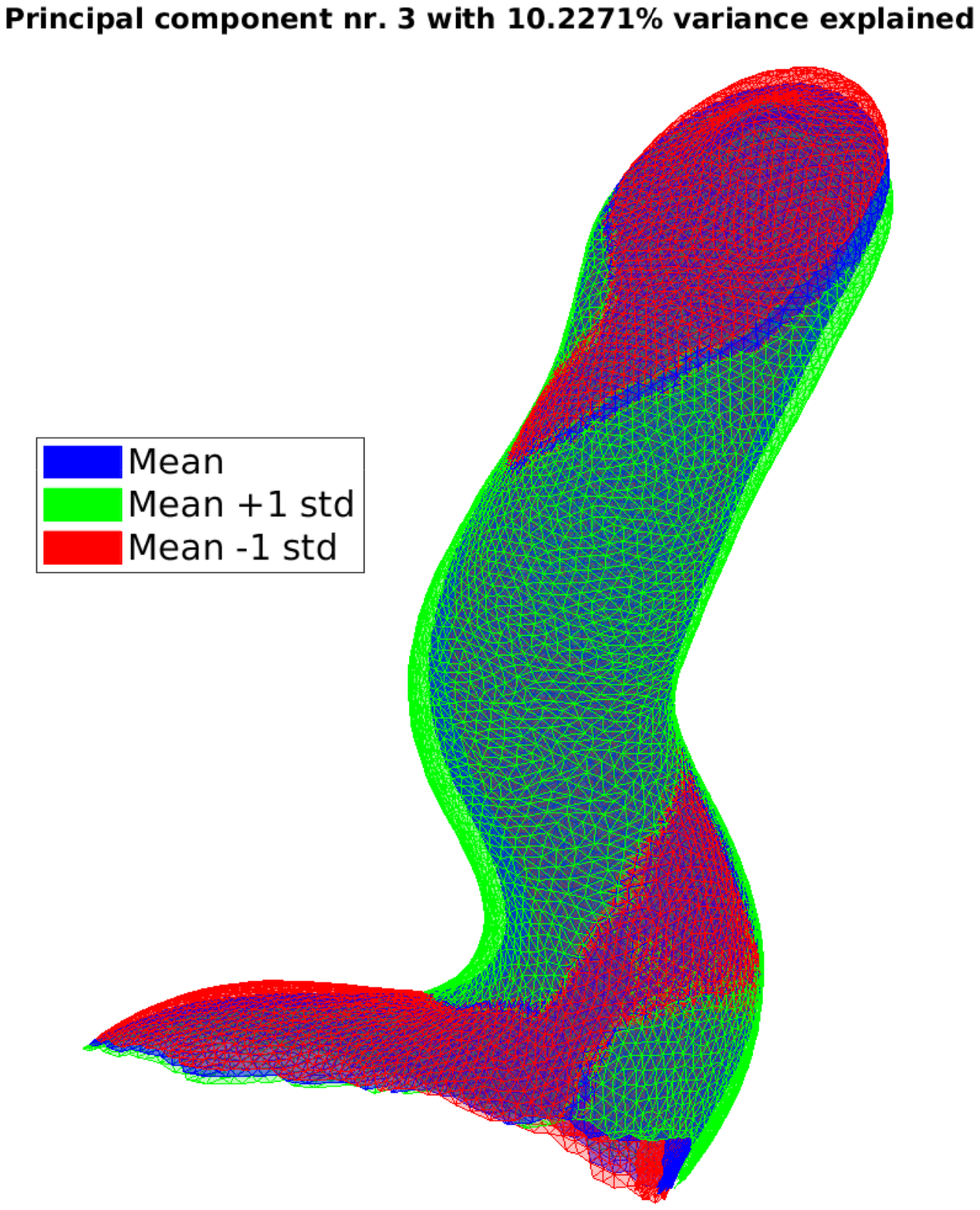}
\caption{PC3 +- 1 std.}
\end{subfigure}
\begin{subfigure}[t]{0.32\textwidth}
\centering
\includegraphics[trim={4cm 4.5cm 4cm 5.2cm},clip,width=1\textwidth]{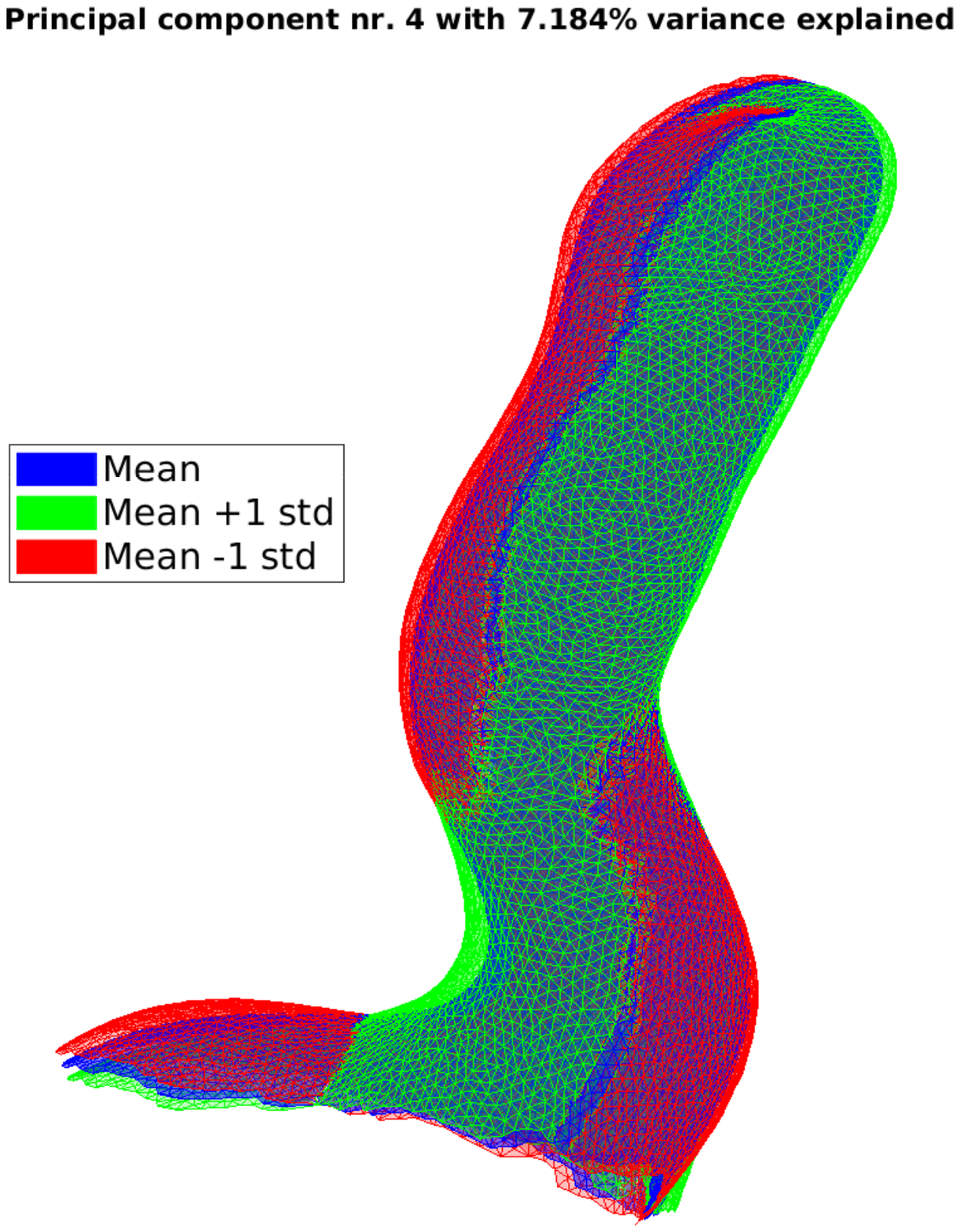}
\caption{PC4 +- 1 std.}
\end{subfigure}
\begin{subfigure}[t]{0.32\textwidth}
\centering
\includegraphics[trim={4cm 4.5cm 4cm 5.2cm},clip,width=1\textwidth]{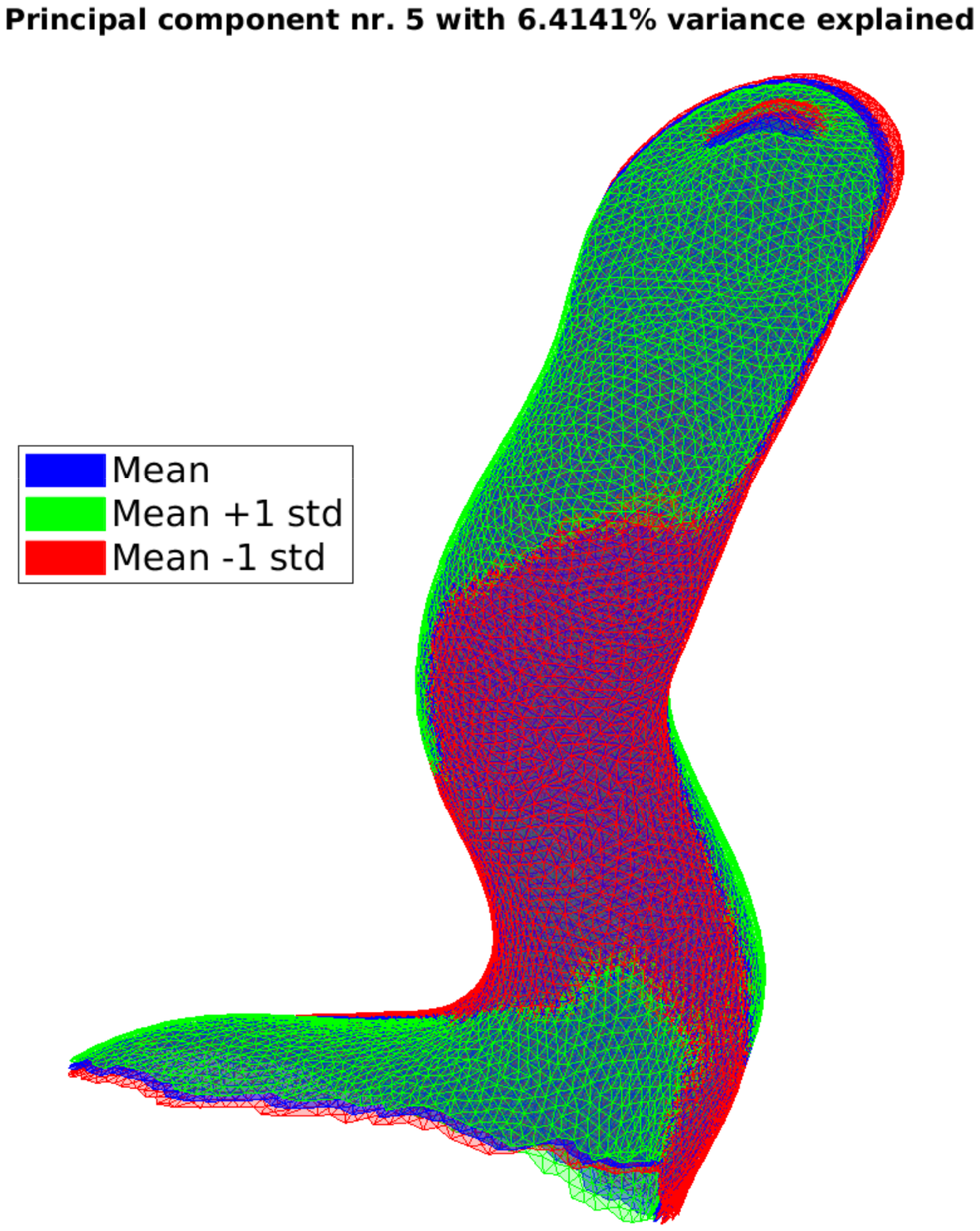}
\caption{PC5 +- 1 std.}
\end{subfigure}
\begin{subfigure}[t]{0.32\textwidth}
\centering
\includegraphics[trim={4cm 4.5cm 4cm 5.2cm},clip,width=1\textwidth]{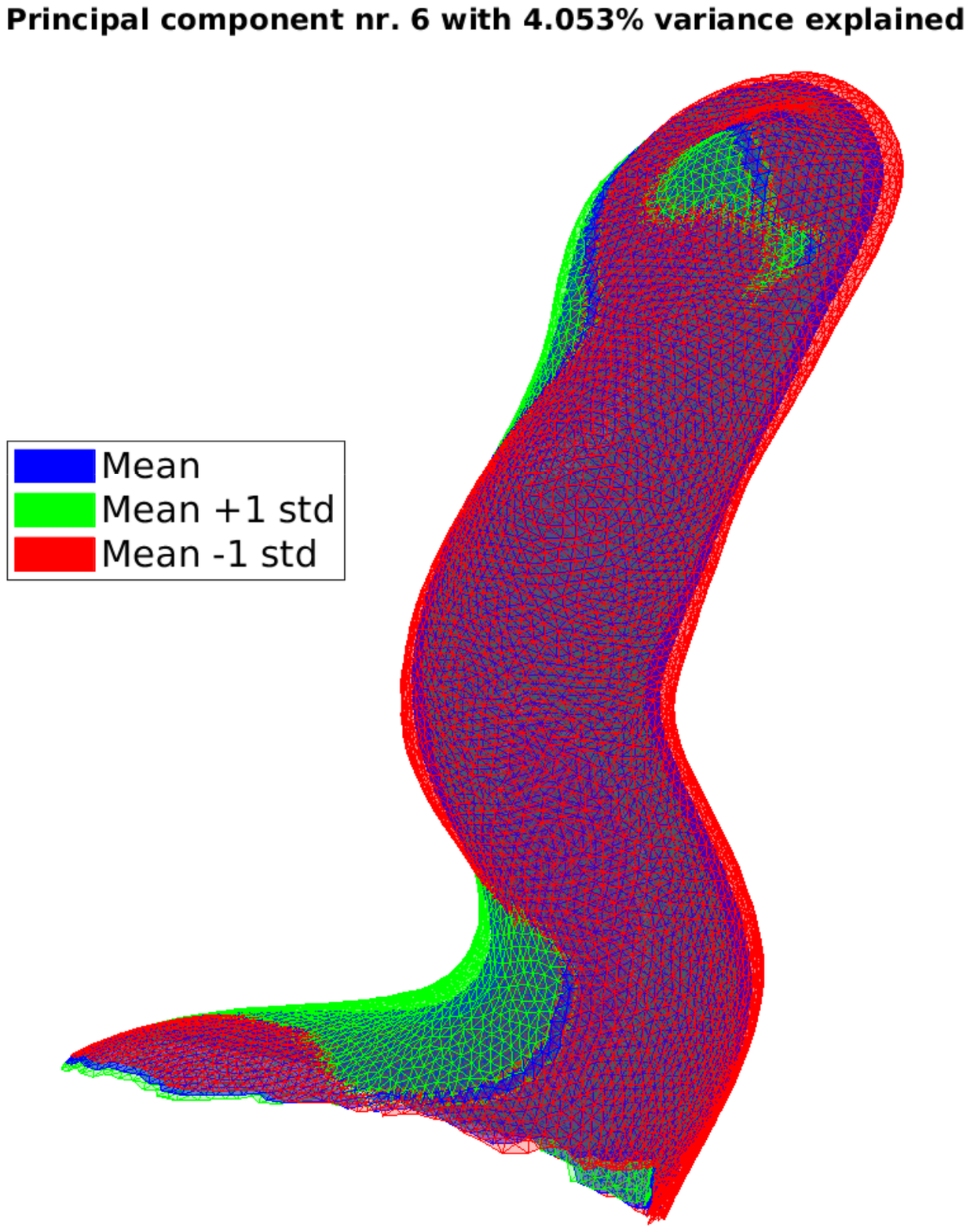}
\caption{PC6 +- 1 std.}
\end{subfigure}
  \end{center}
  \caption{The figure shows the first 6 modes of variations around the mean shape (principal components) explaining roughly 80\% of the total variance in the data set. The blue surface is the mean canal and the red and the green surfaces are the surfaces of the mean +/- 1 standard deviation respectively.  \label{fig:var}}
\end{figure}

The average geometry generated from the 44 subjects has all the anatomical hallmarks of a human ear canal, see \Cref{fig:mean}. The curves, such as the first and second bend can be identified and are similar to that of a real ear canal. Furthermore, the attachment point of the Malleus on the tympanic membrane is easily identified. The slant of the typanic membrane in relation to the canal is also identical to that of an impression of a real ear. 
In addition the variance of the population is very low. \Cref{fig:var} shows the first 6 principal  components (PC) +/- 1 standard deviation. Together the components explain more than 80\% of the total variance in the shape.

\subsection{Acoustical validation}
\begin{figure}
 \includegraphics[width=0.92\linewidth]{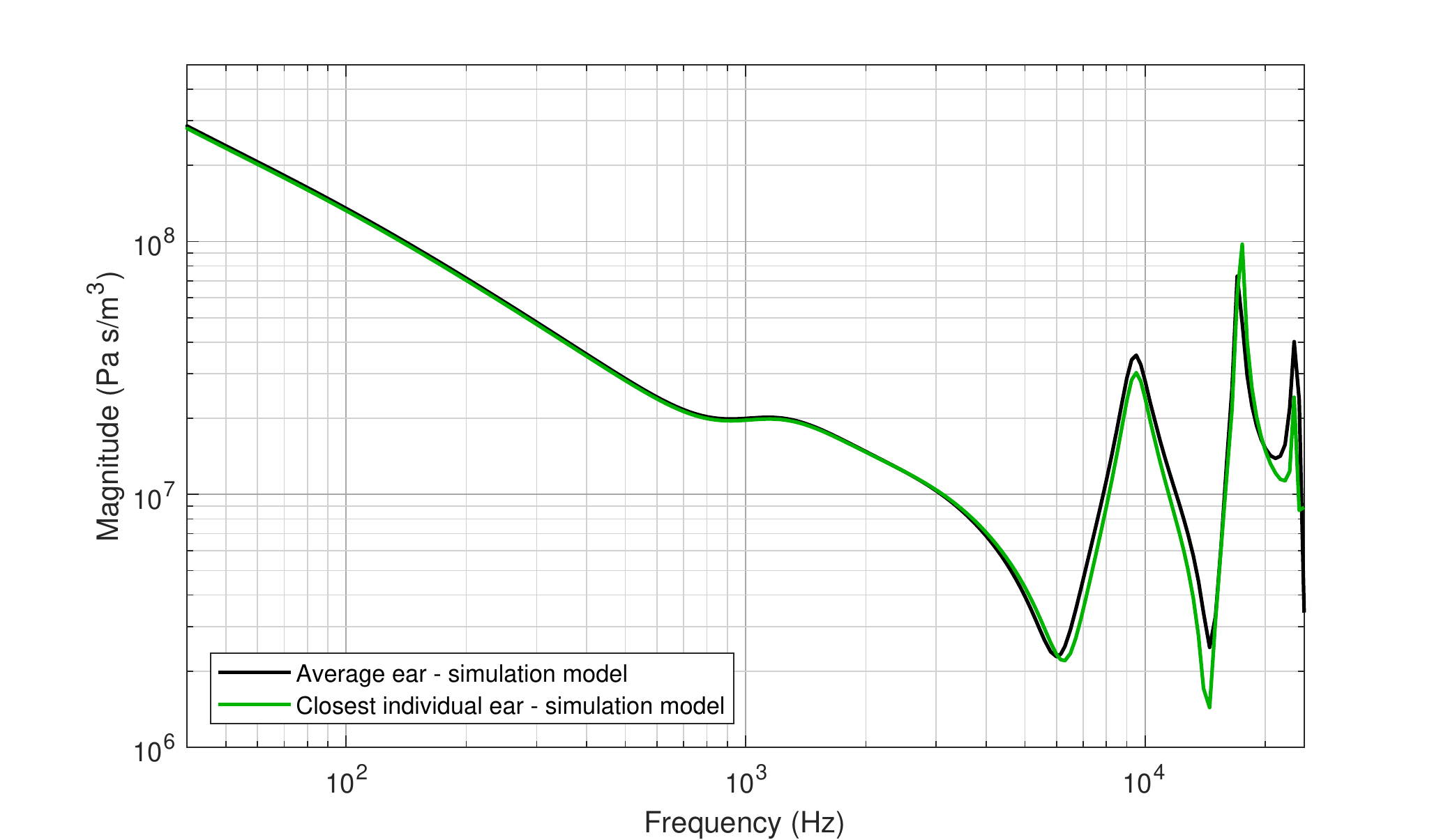}
 \caption{\label{fig:close_06}Magnitude of the impedance output obtained from the simulation model. The bold black curve is the impedance when the average ear canal shape is used in the simulation model and the green curve is the impedance when the individual ear canal closest to the average ear canal shape is used in the model.}
 \end{figure}

\begin{figure}
 \includegraphics[width=0.99\linewidth]{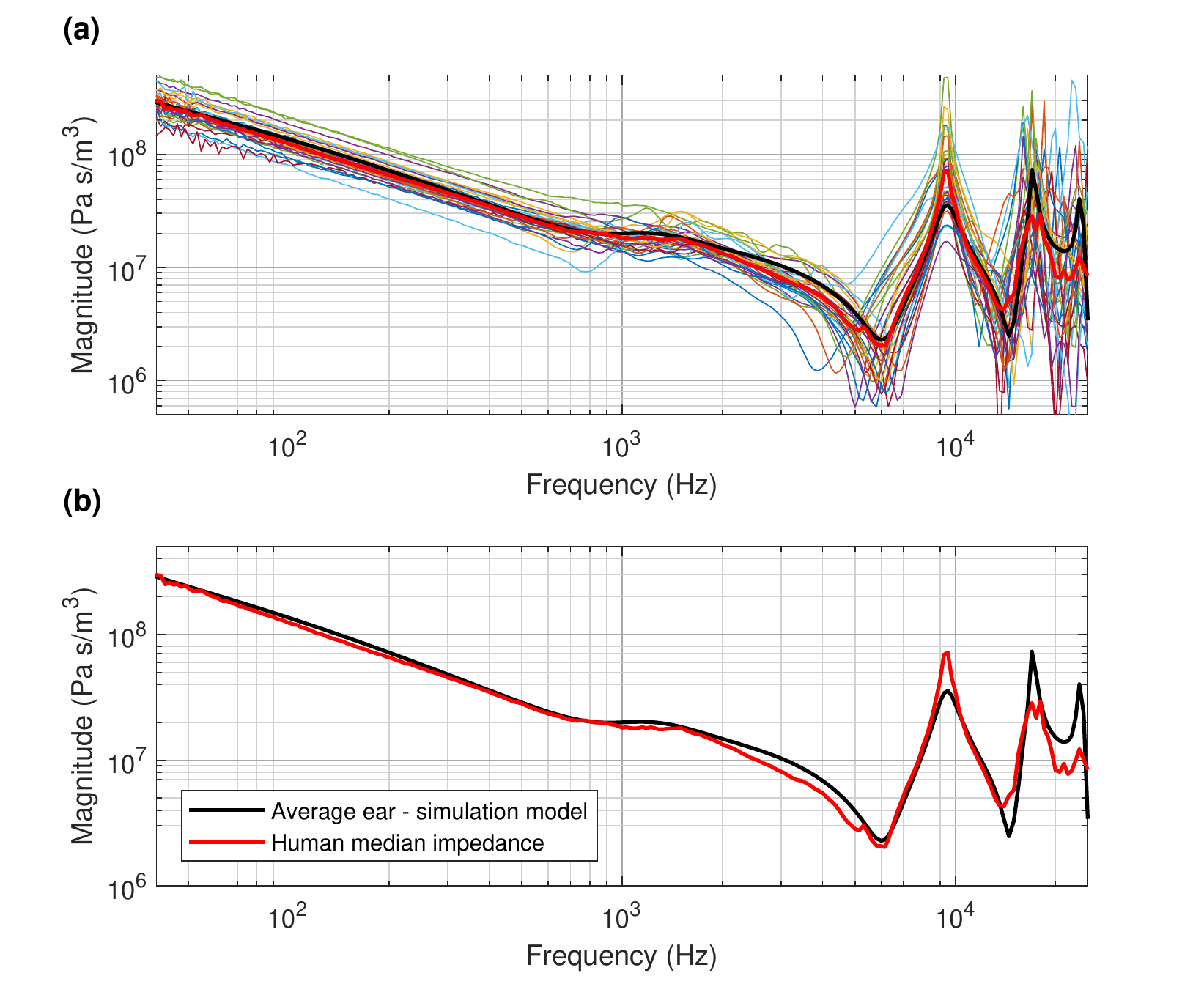}
 \caption{\label{fig:close_median}The magnitude of the propagated measured impedance of the population including the calculated median (the bold red curve also shown in the lower graph) compared to the impedance obtained from the simulation model of the average ear canal (the bold black curve also show in the lower graph).}
 \end{figure}

To validate that the average created not only resembles the shape and anatomical features of a human ear canal, but also bears acoustical properties similar to that of a human ear, we created the average acoustical properties of the population. When we compared that to an acoustic simulation model of the average ear canal geometry imposed by the average human ear drum impedance at the Tympanic membrane. This result can be seen in   \Cref{fig:close_median}. We have also selected the individual ear canal closest to the average shape, see \Cref{fig:close_06} and performed a simulation on this model in place of the average for comparison. This result can be seen in~\Cref{fig:sim}.


\section{Discussion}
The results show that we created an anatomically realistic average human ear canal which not only qualitatively resembles that of a real ear accurately, but also provides a good represenation of the collection of ear canal shapes in the sense that the surfaces exhibit only minor variation around the average ear canal. \Cref{fig:var} shows the first 6 principal components of the shape variation which explains 80\% of the variation in the population and indicates that the population is very homogeneous in spite of the gender composition and age variation. 
The acoustical measurements show that the average impedance of the human impedance measurements corresponds very well to a result of a simulation on our model of the average geometry of the human ear canal and drum impedance. Comparing the simulated impedance of the average ear canal to the most similar ear canal reveals that the two ear canals have almost identical impedance responses. This is a strong indicator that the geometry of the average ear canal does resemble that of a real ear canal. This leads us to conclude that not only is the ear anatomically realistic, the acoustical pattern also fits the distribution of impedance responses similar to that of the real anatomy.

\section{Conclusion}
We have presented the methodology for producing an average ear canal geometry, and applied this to capture the average ear canal geometry of 44 subjects. We have validated that the acoustical properties of this new average ear canal is almost identical to that of the true human average of the impedance responses of the population and to the most similar ear canal. Furthermore, we have shown that the population shape variation around the average ear canal is small. These facts together shows that the average ear canal presented corresponds well to the anatomy of the human ear, that it is a good representation of the population from which it was built, and that it has acoustical properties that are in correspondence with the average acoustical properties of the population. This is to the best of our knowledge the first fully validated shape model of the entire ear canal, and the first acoustically validated average human ear canal.





\bibliographystyle{model1-num-names}




 


\end{document}